\documentclass[10pt,twocolumn,letterpaper]{article}

\usepackage{cvpr}              

\newcommand*{\ShowNotes}{}
\usepackage{algorithm}
\usepackage{algorithmicx}
\usepackage{algpseudocode}
\usepackage{color}
\usepackage{soul}
\usepackage{multirow}
\usepackage{xcolor}
\usepackage{colortbl}
\usepackage{wrapfig}

\newcommand{\xmark}{\ding{55}}%

\definecolor{darkred}{rgb}{0.7,0.1,0.1}
\definecolor{darkgreen}{rgb}{0.1,0.7,0.1}
\definecolor{cyan}{rgb}{0.7,0.0,0.7}
\definecolor{dblue}{rgb}{0.2,0.2,0.8}
\definecolor{maroon}{rgb}{0.76,.13,.28}
\definecolor{burntorange}{rgb}{0.81,.33,0}
\definecolor{tealblue}{rgb}{0.212,0.459, 0.533}
\definecolor{myyellow}{rgb}{0.8627451 , 0.75294118, 0.20784314]}

\definecolor{mypink}{rgb}{0.93359375, 0.62109375, 0.83984375}

\definecolor{pp}{rgb}{0.43921569, 0.18823529, 0.62745098}
\definecolor{rr}{rgb}{0.5254902 , 0.00784314, 0.12941176}
\definecolor{bb}{rgb}{0.09019608, 0.23529412, 0.37647059}
\definecolor{yy}{rgb}{0.49803922, 0.3372549 , 0.0}
\definecolor{gg}{rgb}{0.02352941, 0.3372549 , 0.17647059}
\definecolor{mybrown}{rgb}{0.87058824, 0.56078431, 0.01960784}
\definecolor{myblue}{rgb}{0.3372549 , 0.70588235, 0.91372549}
\definecolor{mypurple}{rgb}{0.8, 0.47058824, 0.7372549 }
\definecolor{myorange}{rgb}{0.835, 0.368, 0}
\definecolor{mygreen}{rgb}{0.00784314, 0.61960784, 0.45098039}
\definecolor{mygt}{rgb}{0.0078125 , 0.57421875, 0.40625}
\definecolor{mysp}{rgb}{0.84765625, 0.515625  , 0.0234375}
\definecolor{mycitecolor}{rgb}{0,0.08,0.45}
\definecolor{mygr}{rgb}{0.9607,0.9607,0.9607}
\definecolor{myoo}{rgb}{0.992,0.9176,0.9019}

\definecolor{myrr}{HTML}{AE031A}
\definecolor{mybb}{HTML}{0155B3}

\definecolor{myteal}{RGB}{128,180,128}
\definecolor{mylavender}{RGB}{150,120,180}

\definecolor{lightgray}{gray}{0.9}
\definecolor{myblue}{rgb}{0.82, 0.88, 0.94}
\definecolor{ours}{rgb}{1.0, 0.93, 0.78} 
\newcommand{\colorours}{\rowcolor{ours}}

\ifdefined\ShowNotes
  \newcommand{\colornote}[3]{{\color{#1}\bf{#2: #3}\normalfont}}
\else
  \newcommand{\colornote}[3]{}
\fi

\usepackage{amsmath,amsfonts,bm}

\def\1{\bm{1}}

\def\sp{space}

\def\vb{{\bm{b}}}

\def\vd{{\bm{d}}}
\def\ve{{\bm{e}}}
\def\vf{{\bm{f}}}

\def\vl{{\bm{l}}}

\def\vp{{\bm{p}}}

\def\vr{{\bm{r}}}
\def\vs{{\bm{s}}}

\DeclareMathAlphabet{\mathsfit}{\encodingdefault}{\sfdefault}{m}{sl}
\SetMathAlphabet{\mathsfit}{bold}{\encodingdefault}{\sfdefault}{bx}{n}

\def\gB{{\mathcal{B}}}

\def\gF{{\mathcal{F}}}
\def\gG{{\mathcal{G}}}

\def\gL{{\mathcal{L}}}
\def\gM{{\mathcal{M}}}

\def\gP{{\mathcal{P}}}

\def\gR{{\mathcal{R}}}

\def\gV{{\mathcal{V}}}

\usepackage{symbols}
\usepackage{pifont}
\usepackage{bbm}
\usepackage{float}
\usepackage{caption}
\usepackage{graphicx}
\usepackage{enumitem}
\usepackage{amssymb}
\usepackage{subcaption}
\usepackage{wrapfig}

\newcommand{\ours}{\texttt{AV3DOD}\@\xspace}
\newcommand{\ourso}{\texttt{AV3DOD} (ours)\xspace}

\newcommand{\TaskName}{Auto-Vocabulary 3D Object Detection }

\definecolor{checkmark}{HTML}{305AFF}
\definecolor{xmark}{HTML}{E62020}
\newcommand{\ccheck}{{\textcolor{checkmark}{\large\checkmark}}}
\newcommand{\ccross}{{\textcolor{xmark}{\large\xmark}}}

\definecolor{cvprblue}{rgb}{0.21,0.49,0.74}
\usepackage[pagebackref,breaklinks,colorlinks,allcolors=cvprblue]{hyperref}

\def\paperID{1327} 
\def\confName{CVPR}
\def\confYear{2026}

\title{Auto-Vocabulary 3D Object Detection}

\author{
    Haomeng Zhang$^{1}$$^\star$  \quad
    Kuan-Chuan Peng$^{2}$ \quad
    Suhas Lohit$^{2}$ \quad
    Raymond A. Yeh$^{1}$ 
    \\
    $^{1}$Department of Computer Science, Purdue University
    \quad\\
    $^{2}$Mitsubishi Electric Research Laboratories (MERL)
    \\
    $^{1}${\tt\small \{zhan5050,rayyeh\}@purdue.edu} \quad
    $^{2}${\tt\small \{kpeng,slohit\}@merl.com}
}

\newcommand\blfootnote[1]{%
  \begingroup
  \renewcommand\thefootnote{}\footnote{#1}%
  \addtocounter{footnote}{-1}%
  \endgroup
}

\begin{document}
\maketitle

\blfootnote{$^\star$ This work was done when Haomeng Zhang was an intern at MERL.}

\begin{abstract}
Open-vocabulary 3D object detection methods are able to localize 3D boxes of classes unseen during training. Despite the name, existing methods rely on user-specified classes both at training and inference. We propose to study {\bf Auto-Vocabulary 3D Object Detection} (AV3DOD), where the classes are automatically generated for the detected objects without any user input. To this end, we introduce Semantic Score (\texttt{SS}) to evaluate the quality of the generated class names. We then develop a novel framework, \ours,
which leverages 2D vision-language models (VLMs) to generate rich semantic candidates through image captioning, pseudo 3D box generation, and feature-space semantics expansion. \ours achieves state-of-the-art (SOTA) performance on both localization (mAP) and semantic quality (\texttt{SS}) on the ScanNetV2 and SUNRGB-D datasets.
Notably, it surpasses the SOTA, CoDA, by 3.48 overall mAP and attains a 24.5\% relative improvement in \texttt{SS} on ScanNetV2.

\end{abstract}
\section{Introduction}
\label{sec:intro}

3D object detection is a fundamental task in computer vision that aims to localize bounding boxes in 3D from a given point cloud.
It serves as the foundation for a wide range of downstream applications, including autonomous driving~\cite{arnold2019survey, mao20233d, wang2020infofocus}, embodied AI~\cite{zhang2024multi, prabhudesai2020embodied, wang2024embodiedscan}, and robotics~\cite{montes2020real, kastner20203d}.
Recent progress in open-vocabulary (OV) learning has enabled 3D detectors to generalize beyond fixed training classes, allowing them to recognize novel objects at test time \cite{cao2023coda, jiao2024unlocking, zhang2024opensight, zeng2023clip2, wang2024ov, cao2025collaborative}.

\begin{figure}[t]
\centering
\includegraphics[width=\linewidth]{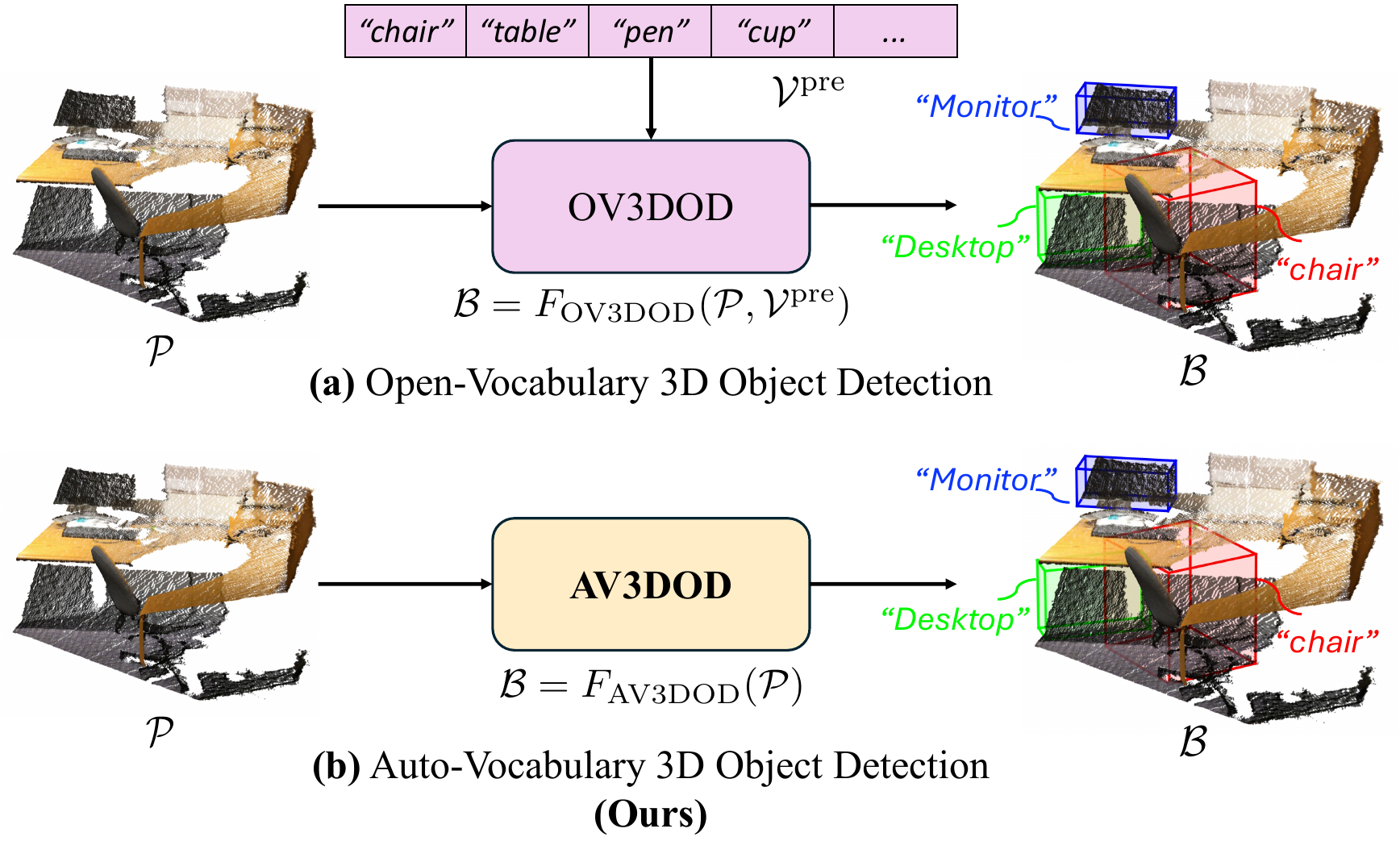}
\caption{
We introduce the task of \textbf{Auto-Vocabulary 3D Object Detection (AV3DOD)}.
\textbf{(a)} Existing Open-Vocabulary 3D Object Detection (OV3DOD) relies on a user-defined vocabulary $\gV^{\text{pre}}$ and the input point cloud $\gP$ to predict 3D objects $\gB$, including bounding boxes and corresponding class labels.
\textbf{(b)} In contrast, AV3DOD needs the model to autonomously derive its own vocabulary and inference directly on $\gP$ without any predefined $\gV^{\text{pre}}$.
}
\label{fig:teaser}
\end{figure}

Despite these advances, existing OV for 3D object detection (OV3DOD) methods remain fundamentally limited.
From the problem formulation perspective, they still \emph{rely on a user-defined vocabulary} at test time, which restricts the scope of semantics and prevents detectors from naming objects beyond the provided classes, \ie, current OV3DOD systems recognize \textit{``unseen''} classes only if their names are already specified in the vocabulary.
This does not fully capture the 
open-world nature of 3D perception, where the detector should be able to localize novel objects and 
\emph{autonomously generate} their class names. 
In terms of semantic discovery, existing methods~\cite{conti2023vocabulary, cao2025collaborative} depend on class-agnostic detectors pre-trained on the base-class annotations.
If the base-trained detector fails to localize a non-base object, no novel class can be discovered downstream.
Moreover, most methods rely solely on explicit textual labels to construct the semantic space, limiting their diversity and ability to represent implicit concepts beyond discrete words.

To address these limitations, we propose the task of \textbf{\TaskName (AV3DOD)}, where a detector must localize 3D objects and jointly discover their class labels \emph{without any user-defined vocabulary}; see comparison of AV3DOD with OV3DOD in~\figref{fig:teaser}.
We further propose a new evaluation metric, \emph{Semantic Score} (\texttt{SS}), which quantifies how well the predicted class labels align with the ground truth, capturing both semantic correctness and detection coverage.
We propose \ours{}, a unified 3D object detection framework that uses 2D vision-language models (VLMs) to mine semantic candidates and provide additional supervision, plus a feature space semantics expansion strategy that samples new semantic prototypes directly in the continuous embedding space, enriching the vocabulary with concepts beyond those expressible by discrete text labels.

We evaluate \ours{} on two 3D object detection benchmarks: ScanNetV2~\cite{dai2017scannet} and SUNRGB-D~\cite{song2015sun}. \ours{} outperforms the open-source SOTA OV3DOD model CoDA~\cite{cao2023coda} and other semantic discovery baselines~\cite{xiao2024florence, yang2025qwen3, misra2021end} in both localization (mAP) and semantic quality (\texttt{SS}) across datasets.
For example, on ScanNetV2, \ours{} beats CoDA by 3.48 overall mAP and achieves a 24.5\% relative improvement in \texttt{SS}, showing its strong ability to accurately localize 3D objects while automatically reasoning about their semantics.

{\noindent\bf Our main contributions are as follows:}
\begin{itemize}
    \item We propose \TaskName (AV3DOD), a new task that extends 3D object detection to an auto-vocabulary setting, where detectors must localize objects and autonomously generate their classes without any user-defined vocabulary.
    \item We propose Semantic Score (\texttt{SS}), a new evaluation metric for joint semantic alignment and detection coverage.
    \item Without annotations, we distill semantic knowledge from 2D VLMs into 3D object representations by generating pseudo 3D object boxes from 2D images.
    \item We design a feature space semantics expansion strategy to capture broader semantics beyond explicit text labels.
\end{itemize}
\section{Related work}
\label{sec:rel}

We compare \ours with the related works in~\tabref{tab:related_works} and elaborate below.

{\bf\noindent Open-vocabulary 2D object detection} (OV2DOD) aims to detect objects with novel classes beyond fixed training labels in 2D images~\cite{xi2025ow, li2024cliff, li2024toward, cheng2024yolo, li2024learning, cho2024language}.
The common approach for the generalization problem is to use large-scale pre-trained visual backbones~\cite{lu2025dynamic, liu2024grounding} or VLMs~\cite{wang2024marvelovd, du2024lami, zhang2025cyclic, zhao2024taming, zhang2024exploring, yao2024detclipv3, kong2024hyperbolic, wang2025declip}.
Other research explores orthogonal gains such as exploiting external knowledge~\cite{li2025benefit}, refining text prompts~\cite{liu2024shine}, introducing negative sampling~\cite{kim2024retrieval}, or generating synthetic data~\cite{feng2024instagen}.
Semantics discovery in 2D is more accessible with powerful 2D models.
In this work, we focus on \emph{auto}-vocabulary \emph{3D} object detection, which is more challenging due to the lack of large-scale pre-trained 3D models and the user-predefined vocabularies.

{\bf\noindent Open-vocabulary 3D object detection} (OV3DOD) extends OV2DOD to 3D scenes, aiming to detect novel objects directly from point clouds.
Most methods build upon the prior closed-set 3D object detection work, such as 3DETR~\cite{misra2021end} (adapted from DETR~\cite{carion2020end}) and VoteNet~\cite{qi2019deep}.
CoDA~\cite{cao2023coda} is the pioneering work that discovers novel objects with class labels unseen during training from 2D in an online manner.
More recently, researchers have explored limited 3D supervision~\cite{zhu2025learning}, training without 3D annotations~\cite{huang2024training, yang2024imov3d, lu2023open, hsu2025openm3d}, transferring knowledge from 2D models~\cite{jiao2024unlocking, zhang2024opensight, zeng2023clip2, wang2024ov, cao2025collaborative}, or using large language models (LLMs)~\cite{peng2024global}. 
While effective, they still need \emph{user-defined vocabularies}.
Moreover, many methods need paired 2D images as auxiliary input during inference~\cite{etchegaray2024find, hsu2025openm3d, huang2024training, xiang2025towards, zhang2024opensight}, restricting their applicability to image-available settings.
In contrast, \ours removes both constraints by enabling the detector to discover and expand its own semantic vocabulary through 2D vision-language guidance during training, while performing inference directly on 3D point clouds without any user-defined vocabulary or 2D inputs.

\begin{table}[t]
\centering
\caption{
\textbf{Characteristic comparison} between \ourso and prior open-vocabulary (OV) 3D object detection/segmentation works (\eg, $\gR_0$: \cite{misra2021end, qi2019deep, xie2020mlcvnet, cheng2021back, zhang2020h3dnet, liu2021group, xie2021venet, zhang2024spatial, huang2024weakly}, $\gR_1$: \cite{cao2023coda, cao2025collaborative, yang2024imov3d, lu2023open, zhu2025learning, jiao2024unlocking, zeng2023clip2, peng2024global, chow2025ov, wang2024ov}, $\gR_2$: \cite{wei20253d}, $\gR_3$: \cite{ma2021delving, oh2024monowad, jiang2024monomae, yan2024monocd, zhang2024decoupled}, $\gR_4$: \cite{etchegaray2024find, hsu2025openm3d, huang2024training, xiang2025towards, zhang2024opensight}, $\gR_5$: \cite{mei2025vocabulary}). Acronym: Vocab.: vocabulary.
}

\setlength{\tabcolsep}{6pt}
\resizebox{\columnwidth}{!}{
\begin{tabular}{lc@{\hspace{3mm}}c@{\hspace{3mm}}c@{\hspace{3mm}}c@{\hspace{3mm}}c@{\hspace{3mm}}c>{\columncolor{ours}}c}
\toprule
\multirow{2.5}{*}{\textbf{\shortstack{Open/Auto-Vocab. 3D Object\\Detection Method Conditions}}} & \multicolumn{7}{c}{\textbf{Method Categories}} \\ 
\cmidrule(l){2-8}
 & $\gR_0$ & $\gR_1$ & $\gR_2$ & $\gR_3$ & $\gR_4$ & $\gR_5$ & \ours \\
\midrule
support 3D object detection & \ccheck & \ccheck & \ccross & \ccheck & \ccheck &  \ccross & \ccheck \\
support OV object detection & \ccross & \ccheck & \ccross & \ccross & \ccheck &  \ccross & \ccheck \\
need \textbf{no} user-defined vocabulary & \ccross & \ccross & \ccheck & \ccross & \ccross  & \ccheck & \ccheck \\
need \textbf{no} 2D images at inference & \ccheck & \ccheck & \ccheck & \ccross & \ccross & \ccross &\ccheck \\
non-discrete vocabulary features & \ccross & \ccross & \ccross & \ccross & \ccross & \ccross & \ccheck \\
\bottomrule
\end{tabular}
}
\label{tab:related_works}
\end{table}

{\bf\noindent Auto-vocabulary (AV) learning.}
Recent efforts in AV learning aim to enable models to construct and adapt vocabularies without explicit human supervision.
This paradigm has been explored in classification~\cite{conti2023vocabulary} and segmentation~\cite{wei20253d, ulger2025auto, conti2024vocabulary, mei2025vocabulary}.
Inspired by this direction, \ours is the first to bring the AV concept into 3D object detection.
Unlike existing AV methods that mainly build vocabularies and rely on a frozen OV expert for inference, \ours jointly mines richer semantics and uses them to improve the detector at training.
Moreover, most OV/AV methods encode semantics with discrete textual labels, limiting their models to a fixed, word-level vocabulary.
In contrast, \ours proposes a feature space semantics expansion strategy, sampling new representations in continuous embedding space, enabling the model to capture and reason about concepts beyond those expressible by explicit text.

\section{Preliminaries}
\label{sec:prelim}

{\bf\noindent Open-vocabulary 3D object detection} (OV3DOD) studies how to detect 3D objects belonging to classes beyond the observed ones during training. At test time, the model is provided with a point cloud $\gP$ and a pre-defined vocabulary $\gV^{\text{pre}}$, which consists of two disjoint subsets: $\gV^{\text{base}}$, denoting the base classes with 3D box annotations available during training, and $\gV^{\text{novel}}$, denoting the unseen novel classes. The 3D detector predicts a set of 3D bounding boxes $\gB = \{\vb_i\}_{i=1}^{N}$, where each box $\vb_i = (c_i, \vl_i)$ consists of the box's geometric representation $\vl_i$ (location, size, and orientation) and a class label $c_i \in \gV^{\text{pre}}$. The detection process can be expressed as:
\be
\label{eq:ov3d}
\gB = F_{\text{OV3DOD}}(\gP, \gV^{\text{pre}}),
\ee
where $\gB$ depends explicitly on $\gV^{\text{pre}}$ given by the user.

{\bf\noindent OV3DOD evaluation.}  
Existing OV3DOD methods~\cite{cao2023coda, misra2021end} evaluate each detected box against the pre-defined classes $\gV^{\text{pre}}$ to obtain per-class predictions, and the standard detection metrics (\eg, mean Average Precision (mAP)) are reported.
This evaluation protocol only measures how well a model aligns with \emph{the user-defined vocabulary}, not its ability to discover or reason about novel classes beyond them.
\section{\TaskName{}}
\label{sec:task}

\subsection{Problem formulation}
Given a point cloud $\gP$, \TaskName (AV3DOD) localizes the objects in the 3D scene while jointly discovering their classes \emph{without relying on any user-specified vocabulary}.
The model predicts a set of 3D bounding boxes $\gB = \{\vb_i\}_{i=1}^{N}$, where each box $\vb_i = (c_i, \vl_i)$ consists of the box's geometric representation $\vl_i$ (location, size, and orientation) and an \emph{automatically discovered} class label $c_i$.
Formally,
\be
\label{eq:av3d}
\gB = F_{\text{AV3DOD}}(\gP).
\ee
Each predicted label $c_i$ is a free-form language noun phrase (\eg, `chair', `studio couch') that can be drawn from any word in the English lexicon, supported by the text decoder in the model.
This differs from the OV3DOD setting in~\equref{eq:ov3d} in that the set of classes needs to be learned as part of the model. At training time, point clouds and their corresponding annotations of 3D boxes in $\gV^{\text{base}}$ are provided. The model is not aware of $\gV^{\text{novel}}$, nor is it given any 3D box annotations from $\gV^{\text{novel}}$. 

\subsection{Evaluating AV3DOD}
\label{sec:eval_protocol}
In OV3DOD, models are evaluated on a fixed, user-defined set of object classes and their corresponding box localizations. However, this evaluation does not account for the quality of the discovered classes needed for AV3DOD. To address this limitation, we propose a new evaluation protocol that jointly measures both the accuracy of class discovery and the quality of 3D box localization.

{\bf\noindent Localization.}
Following~\cite{cao2023coda}, we report mAP and Average Recall (AR) for all, base, and novel classes.
Since mAP is defined per class, we first assign each predicted class with the closest ground truth class by measuring text similarity using  CLIP~\cite{radford2021learning}.
With the assignment, the per-class AP values are computed and averaged to obtain mAP.
This assignment ensures that semantically related predictions (\eg, `sofa' \vs `couch') are appropriately credited, allowing consistent comparison even when models generate class labels that do not exactly match the ground-truth vocabulary.

{\bf\noindent Semantics.} To evaluate the quality of predicted classes, we introduce a new \emph{semantic score} (\texttt{SS}) that measures how well a predicted class is aligned to the ground-truth class in the embedding space. 
For each ground-truth box, we identify its matched prediction as the one with the highest Intersection-over-Union (IoU) exceeding a predefined threshold.
If no prediction meets this criterion, the ground-truth box is considered unmatched and excluded from the evaluation.
For each matched pair, we compute the cosine similarity between the predicted and ground-truth class embeddings in the CLIP~\cite{radford2021learning} text feature space.
A prediction is considered semantically correct if its similarity is above a pre-defined threshold.
By varying the threshold, we compute the area under the curve (AUC) for base and novel classes separately, denoted as $\operatorname{AUC}_{\text{b}}$ and $\operatorname{AUC}_{\text{n}}$.  

Since AUC is computed only over matched pairs, it reflects semantic alignment conditioned on successful detections.
Hence, a model that detects only a small subset of objects, such as confident base-class instances, may still obtain a high AUC despite poor overall coverage.
To address this, we weight each AUC by the \emph{coverage} ($\operatorname{Cov}$), which measures the proportion of ground-truth boxes correctly detected by the model, \ie, having at least one predicted box with an IoU above a threshold.
The coverage for base and novel classes is denoted as $\operatorname{Cov}_{\text{b}}$ and $\operatorname{Cov}_{\text{n}}$.
We then compute separate semantic scores for base and novel classes:
\be
\texttt{SS}_\text{b} = \operatorname{AUC}_{\text{b}}\times \operatorname{Cov}_{\text{b}}, \quad 
\texttt{SS}_\text{n} = \operatorname{AUC}_{\text{n}}\times \operatorname{Cov}_{\text{n}}
\ee
The final semantic score \texttt{SS} aggregates these two scores by weighting them according to the proportion of ground-truth boxes from each class type:
\be
\label{eq:ss}
\texttt{SS} = \alpha_\text{b} \times \texttt{SS}_\text{b} + \alpha_\text{n} \times \texttt{SS}_\text{n},
\ee
where $\alpha_{\text{b}}$/$\alpha_{\text{n}}$ is the relative proportions of the base/novel class objects.
More evaluation details are in the supplement.

\begin{figure*}[t]
\centering
\includegraphics[width=0.98\linewidth]{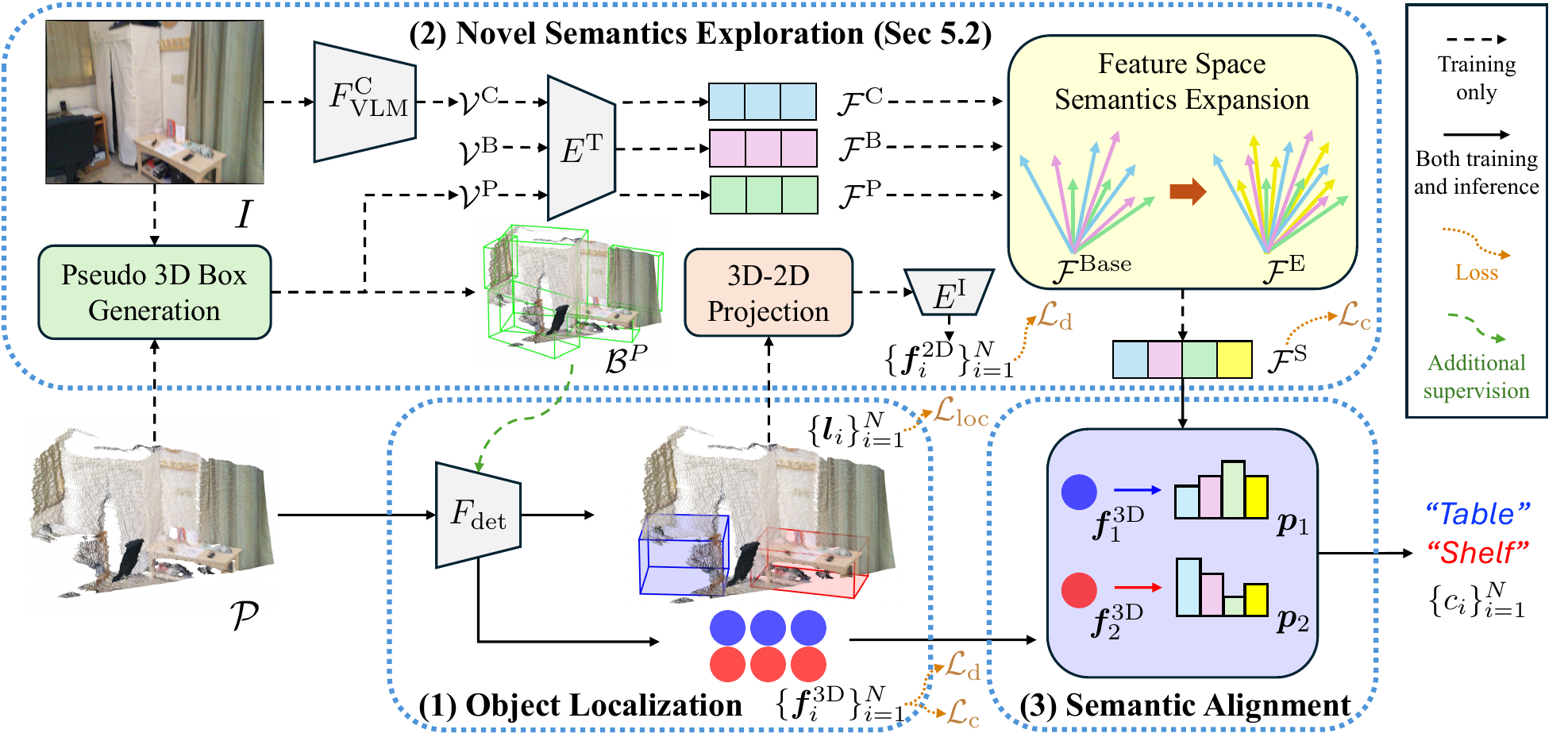}
\caption{
\textbf{Illustration of \ours (\secref{sec:overview}).}
\ours\ consists of three key components:
(1) \emph{Object Localization}, which uses $F_{\text{det}}$ to generate class-agnostic 3D object proposals $\{\vl_i, \vf_i^{3D}\}_{i=1}^N$ from $\gP$;
(2) \emph{Novel Semantics Exploration} (\secref{sec:semantic_exploration}), which constructs $\gF^{\text S}$ by integrating base class features $\gF^{\text B}$, VLM caption features $\gF^{\text C}$, pseudo label features $\gF^{\text P}$, and expanded features $\gF^{\text E}$; and
(3) \emph{Semantic Alignment}, which aligns the detected 3D object features with $\gF^{\text S}$ to predict class labels $\{c_i\}_{i=1}^N$.
Input $\gP$ is colorized for visualization.
}
\label{fig:pipeline}
\end{figure*}
\section{\ours{}}
\label{sec:approach}

\subsection{Overview}
\label{sec:overview}

We propose \ours, a novel framework for the \TaskName task that uses 2D VLMs to discover semantic candidates and provide additional supervision for 3D object detection.
\ours comprises three key components:
(1) an object localization module that generates class-agnostic 3D object proposals and corresponding object features; 
(2) a novel semantics exploration module that constructs an enriched vocabulary in the feature space by integrating semantics from multiple sources; and
(3) a semantics alignment module that matches detected 3D object features with the semantic features to predict class labels. 
An overview of \ours is illustrated in \figref{fig:pipeline}.

{\bf\noindent The object localization} module $F_{\text{det}}$ produces class-agnostic 3D object proposals and the corresponding object features from an input point cloud $\gP$:
\be
F_{\text{det}}(\gP) = \{\vd_i\}_{i=1}^{N},
\ee
where each detection $\vd_i = (\vl_i, \vf_i^{3D})$ consists of the 3D box geometric parameters $\vl_i$ and its 3D feature $\vf_i^{3D}$.
We use 3DETR~\cite{misra2021end}, pre-trained on base-class box annotations, as the backbone detector and fine-tune it during training.

{\bf\noindent The novel semantics exploration} module aims to construct a set of super vocabulary features $\gF^{\text S}$ that captures rich semantic candidates beyond the known base vocabulary $\gV^B$.
These generated features are used to predict the class label for each detected 3D object proposal.
We construct $\gF^{\text S}$ by integrating four complementary sources: 
(i) base class features $\gF^{\text B}$ from the base class vocabulary;
(ii) VLM caption features $\gF^{\text C}$ generated from VLM-generated captions;
(iii) pseudo label features $\gF^{\text P}$ from pseudo 3D boxes; and
(iv) expanded features $\gF^{\text E}$ obtained through feature space semantics expansion.
The final unified $\gF^{\text{S}}$ is given by:
\be
\gF^{\text{S}} = \gF^{\text B} \cup \gF^{\text C} \cup \gF^{\text P} \cup \gF^{\text E}.
\ee
We detail the novel semantics exploration in \secref{sec:semantic_exploration}.

{\bf\noindent The semantic alignment} module predicts class labels for detected 3D object proposals by matching their 3D object features $\vf_i^{3D}$ with the super vocabulary features $\gF^{\text S}$.
To bridge the modality gap between 3D and text representations, we follow~\citet{cao2023coda} and distill the corresponding 2D CLIP image features into 3D object features using the loss defined in~\equref{eq:2d_distill}.
The class probability for each proposal $\vd_i$ is then computed via normalized similarity:
\be
\vp_{i, k} = \frac{\operatorname{exp}(\vf_i^{3D} \cdot \vf^s_{k})}{\sum_{q=1}^{|\gF^{\text{S}}|} \operatorname{exp}(\vf_i^{3D} \cdot \vf^s_{q})},
\ee
where `$\cdot$' denotes the dot product, and $\vf^s_{k}$ is the feature embedding of class $c^k$ in $\gF^{\text S}$.
The final class label of each 3D object proposal is decided by:
\be
c_i = \operatorname*{argmax}_{\gF^{\text{S}}}(\vp_{i, k}). 
\ee

\begin{figure}[t]
\centering
\includegraphics[width=\linewidth]{raw/pseudo_box.pdf}
\caption{
\textbf{Illustration of pseudo 3D box generation.}
A 2D VLM produces an object segmentation mask $M_k^P$ and a textual label $c_k^P$ for each detected object.
The point cloud $\gP$ is projected onto the image $I$ to extract the corresponding point group $\gG^P_k$, from which a pseudo 3D bounding box is estimated via PCA.
Input $\gP$ is colorized for visualization.
}
\label{fig:pseudo_box_generation}
\end{figure}

\subsection{Novel semantics exploration}
\label{sec:semantic_exploration}
{\bf\noindent Semantics discovery from VLM captions.}
To enrich the semantic space beyond the base class vocabulary, we use the captioning capability of existing 2D VLMs to automatically extract diverse object classes from training images.
For each training image $I_{i}$, we query the 2D VLM Florence2~\cite{xiao2024florence} ($F_{\text{VLM}}^{\text C}$) to generate a descriptive caption $C_i$.
We then apply the NLP toolkit spaCy \cite{spaCy} to parse each caption and extract all noun tokens from $C_i$, forming an image-specific vocabulary $\gV_i$.
Aggregating across all training images yields the caption-derived vocabulary $\gV^\text{C} = \bigcup_{i} \gV_{i}$.
Each word in $\gV^\text{C}$ is encoded by the text encoder $E^{\text{T}}$ to construct VLM caption features $\gF^\text{C}$:
\be
\gF^\text{C} = \{\, \vf^{C}_j = E^{\text{T}}(c_j) \mid c_j \in \gV^\text{C} \,\}.
\ee

{\bf\noindent 3D pseudo box generation via 2D VLM.}
\citet{cao2023coda} could only discover novel object classes from 3D detections produced by a pre-trained detector, which largely limits semantic diversity.  
To overcome this limitation, we propose a pseudo 3D box generation framework (\figref{fig:pseudo_box_generation}) that creates additional 3D object proposals with corresponding semantic labels inferred from a 2D VLM.
These pseudo 3D boxes provide supplementary training signals that enrich the vocabulary and improve 3D localization, especially for categories beyond the annotated base classes.

For each training image $I$, we query the 2D VLM Florence2~\cite{xiao2024florence} that detects $K$ objects in the scene:
\be
\{(M^P_k, c^P_k)\}_{k=1}^{K} = F_{\text{VLM}}^{\text D}(I),
\ee
where $M_k^P$ is the 2D object mask and $c_k^P$ is its predicted semantic label.
All 3D points $\gP$ are projected onto $I$ and aligned with the 2D object masks to obtain a 3D point mask:
\be
\gM^P_k = \{\,\vp \in \gP \mid \pi(\vp, I) \in M^P_k \,\},
\ee
where $\pi(\cdot)$ denotes the 3D-to-2D projection.
For each 3D mask $\gM_k$, we estimate the object orientation and size by applying principal component analysis (PCA)~\cite{jolliffe2011principal} on the $x$–$y$ coordinates.
The final oriented 3D bounding box is represented by its geometric parameters $\vl^{P}_k$, and the pseudo 3D object $\vb^{P}_k = (c^{P}_{k}, \vl^{P}_{k})$.
All generated pseudo labels from the training set form the pseudo box set $\gB^{P} = \{\vb^{P}_k\}_{k=1}^{N_P}$, and their unique textual labels form the pseudo vocabulary:
\be
\gV^\text{P} = \{c^{P}_{k} \mid \vb^{P}_k \in \gB^{P}\}.
\ee
Each word in $\gV^\text{P}$ is encoded by the text encoder $E^{\text{T}}$ to construct pseudo label features $\gF^\text{P}$.

{\bf\noindent Feature space semantics expansion (FSSE).}
To broaden the semantic representation beyond explicit textual concepts, we propose the FSSE strategy.
Unlike prior modules that construct the vocabulary via discrete vocabulary terms, FSSE directly generates new semantic prototypes by sampling \textit{in the continuous embedding space}.

We first construct the base feature set $\gF^{\text{Base}}$ by concatenating vocabulary features from base class features $\gF^\text{B}$, VLM caption features $\gF^\text{C}$, and pseudo label features $\gF^\text{P}$.
A reference feature $\vr \in \gF^{\text{Base}}$ is selected uniformly at random.
We then perturb it along a random normalized direction $\vd$ drawn from a multivariate standard normal distribution, \ie, $\vd \sim \mathcal{N}(0, I_D)$, where $D$ denotes the feature dimension and $I_D$ is the $D \times D$ identity matrix.
The perturbation magnitude is controlled by a scalar $\lambda \sim \mathcal{U}(0,1)$, generating a candidate expanded feature $\ve$:
\be
\ve = \operatorname{norm}(\vr + \lambda \vd).
\ee
To avoid outliers, we impose a lower bound $\theta^{E}_{\text{min}}$ on the Cosine similarity between $\ve$ and $\vr$:
\be
\label{eq:lower}
\operatorname{Cos}(\vr, \ve) > \theta^{E}_{\text{min}}
\ee
To ensure the diversity of the expanded features, each $\ve$ must be sufficiently dissimilar to existing sampled expanded feature in $\gF^{\text{E}}$ and the base features in $\gF^{\text{Base}}$. We enforce this with a threshold on the cosine similarity:
\begin{align}
\label{eq:diverse}
\max_{\vf \in (\gF^{\text{Base}} \cup \gF^\text{E})} \operatorname{Cos}(\vf, \ve) &< \theta^{E}_{\text{max}},
\end{align}
Only the candidates that satisfy~\equref{eq:lower} and ~\equref{eq:diverse} are added to $\gF^\text{E}$. We repeat this process until we obtain $K$ valid samples.
We detail FSSE in~\alggref{alg:sampling}.

\begin{algorithm}[t]
\caption{Feature Space Semantics Expansion (FSSE)}
\label{alg:sampling}
\begin{algorithmic}[1]
\State \textbf{Inputs:} Base features $\gF^{\text{Base}}$, number of samples $K$, similarity range $[\theta^{E}_{\min}, \theta^{E}_{\max}]$.
\State \textbf{Outputs:} Expanded features $\gF^\text{E}$.
\State $\gF^\text{E} \leftarrow \emptyset$
\While{$|\gF^\text{E}| < K$}
    \State $\vr \leftarrow \operatorname{RandomSample}(\mathbf{B})$
    \State $\vd \sim \mathcal{N}(0, I_D)$; $\mathbf{d} \leftarrow \vd/\|\vd\|$
    \State $\lambda \sim \mathcal{U}(0,1)$
    \State $\ve \leftarrow \text{norm}(\vr + \lambda\mathbf{d})$
    \State {\color{gray}\# Check similarity to existing features.}
    \State $s_{\text{r}} \leftarrow \operatorname{Cos}(\vr, \ve)$
    \State $s_{\text{f}} \leftarrow \max_{\vf \in (\gF^{\text{Base}} \cup \gF^\text{E})} \operatorname{Cos}(\vf, \ve)$
    \If{$s_{\text{r}} > \theta^{E}_{\min}$ \textbf{and} $s_{\text{f}} < \theta^{E}_{\max}$}
        \State $\gF^\text{E} \leftarrow \gF^\text{E} \cup \{\ve\}$
    \EndIf
\EndWhile
\State \textbf{Return:} $\gF^\text{E}$
\end{algorithmic}
\vspace{2pt}
{\raggedright\footnotesize $\operatorname{Cos}(\mathbf{x}, \mathbf{y}) = \frac{\mathbf{x}^\top \mathbf{y}}{\|\mathbf{x}\|\,\|\mathbf{y}\|}$ denotes cosine similarity between features.\par}
\end{algorithm}

\begin{figure}[t]
\centering
\includegraphics[width=\linewidth]{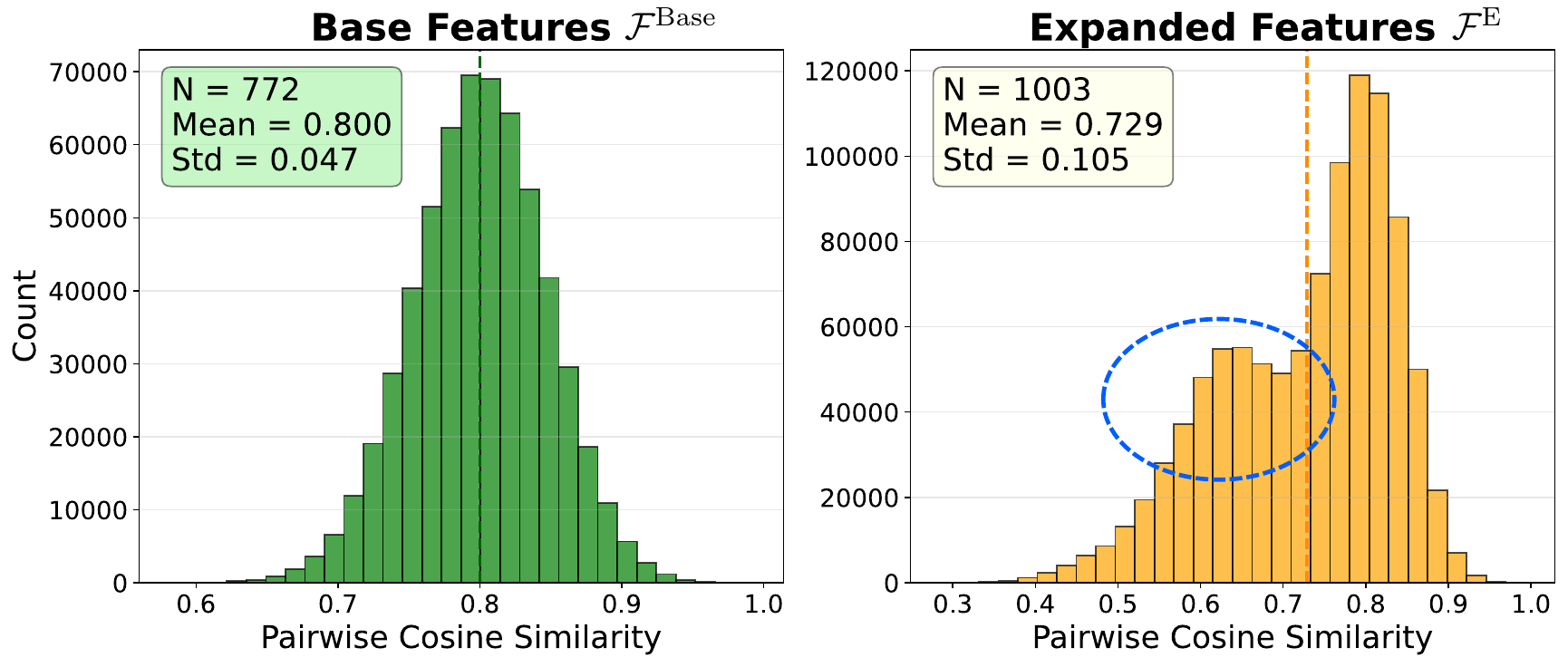}
\caption{
\textbf{Impact of feature space semantics expansion.}
We compare the distribution of pairwise cosine similarity between vocabulary features before and after expansion.
The expanded feature space is generated by sampling 30\% extra features, with similarity thresholds $\theta^{E}_{\text{min}} = 0.6$ and $\theta^{E}_{\text{max}} = 0.9$.
The \textcolor{blue}{highlighted region} on the right demonstrates the newly introduced diverse feature pairs, indicating broader semantics after expansion.
}
\label{fig:feature_expansion}
\end{figure}

We visualize the impact of FSSE in \figref{fig:feature_expansion}.
The highlighted region in the $\gF^\text{E}$ distribution illustrates newly sampled, diverse feature pairs introduced by the expansion process.
Compared with $\gF^{\text{Base}}$, $\gF^\text{E}$ shows an 8.92\% relative decrease in mean pairwise cosine similarity and a 123.06\% increase in standard deviation, which shows that our FSSE effectively broadens and diversifies the semantic space.

\subsection{Training objective}
\label{sec:objective}

{\bf\noindent Localization losses.}
For object localization, we adopt the same objectives and balancing weights as 3DETR~\cite{misra2021end}, including the Hungarian matching-based losses for object-ness, bounding box regression, and orientation prediction:
\fontsize{9.6pt}{10pt}
\be
\label{eq:loss_loc}
\gL_{\text{loc}} = \lambda_{\text{obj}}\gL_{\text{obj}} + \lambda_{\text{center}}\gL_{\text{center}} + \lambda_{\text{size}}\gL_{\text{size}} + \lambda_{\text{angle}}\gL_{\text{angle}}.
\ee
\normalsize
During training, both annotated ground-truth 3D boxes for base classes and pseudo 3D boxes generated from 2D VLM detections are treated as supervision targets.
This ensures predicted boxes align with ground-truth annotations or pseudo 3D proposals in position and geometry.

{\bf\noindent Semantics losses.}
For semantics, we follow~\citet{cao2023coda} to include both a 3D-2D distillation loss $\gL_{\text{d}}$ and a 3D-Text contrastive alignment loss $\gL_{\text{c}}$.
$\gL_{\text{d}}$ is defined as:
\be
\label{eq:2d_distill}
\gL_{\text{d}} = \frac{1}{N}\sum_{i=1}^{N} \norm{\vf_i^{3D} - \vf_i^{2D}}_1,
\ee
where $\vf_i^{3D}$ is the 3D object feature and $\vf_i^{2D}$ is the associated 2D CLIP image feature.
To obtain $\vf_i^{2D}$, we project the predicted 3D box onto $I$ and encode the resulting region with the CLIP image encoder $E^{\text{I}}$.
$\gL_{\text{d}}$ distills the pre-trained 2D semantics into 3D object features, helping the alignment to the CLIP text feature space.
$\gL_{\text{c}}$ is formulated as:
\be
\gL_{\text{c}} 
= -\frac{1}{N} \sum_{i=1}^{N}
\log 
\frac{
\exp(\vf_i^{3D} \cdot \vf_{y_i}^{s} / \tau)
}{
\sum_{k=1}^{|\gF^{\text{S}}|} \exp(\vf_i^{3D} \cdot \vf_k^{s} / \tau)
},
\ee
where $y_i$ is the class label for the $i^{\text{th}}$ object and $\tau$ is a temperature coefficient.
$\gL_{\text{c}}$ encourages strong semantic alignment between 3D object features and text features.

{\bf\noindent Overall objective.}
The final training loss $\gL_{\text{total}}$ integrates both localization and semantic objectives, jointly optimized for both annotated and pseudo boxes:
\be
\label{eq:loss_final}
\gL_{\text{total}} = \gL_{\text{loc}} + \lambda_{\text{d}} \gL_{\text{d}} + \lambda_{\text{c}} \gL_{\text{c}},
\ee
where $\lambda_{\text{d}}$ and $\lambda_{\text{c}}$ are the corresponding loss weights.
\section{Experiments}
\label{sec:exp}

\begin{table*}[t]
\centering
\caption{
\textbf{Quantitative comparison on ScanNet dataset.}
For all metrics, higher is better. The best and second best performances are marked in \textbf{bold} and \ul{underline}. Acronyms: mAP: mean Average Precision; Rec: Recall; AUC: Area Under Curve; Cov: Coverage; \texttt{SS}: Semantic Score. The subscript `b'/`n' denotes `base'/`novel' classes.
}
\resizebox{1\linewidth}{!}{
    \begin{tabular}{lccccccc|ccccccc}
    \specialrule{.15em}{.05em}{.05em}
    &  & \multicolumn{6}{c}{Localization} & \multicolumn{7}{c}{Semantics} \\
    \cline{3-15} 
    \multirow{-2}{*}{Method} & \multirow{-2}{*}{Setting} & mAP & mAP$_\text{b}$ & mAP$_\text{n}$ & Rec & Rec$_\text{b}$ & Rec$_\text{n}$ & AUC$_\text{b}$ & Cov$_\text{b}$ & AUC$_\text{n}$ & Cov$_\text{n}$ & \texttt{SS}$_\text{b}$ & \texttt{SS}$_\text{n}$ & \texttt{SS} \\ 
    \hline
    3DETR+PointLLM & zero-shot & 2.84 & 14.89 & 0.43 & 6.88 & 32.77 & 1.71 & \ul{0.896} & 0.616 & 0.754 & 0.369 &0.552 &0.278 & 0.389 \\
    3DETR+Qwen3 & zero-shot & 3.18 & 11.85 & 3.18 & 9.93 & 22.59 & 7.39 & 0.887 & 0.618 & 0.749 & 0.369 &0.548 &0.276 &0.387 \\
    3DETR+Florence2 & zero-shot & 2.83 & 13.82 & 0.63 & 8.06 & 28.00 & 4.07 & 0.867 & 0.616 & 0.749 & 0.371 &0.534 &0.278 & 0.382 \\
    3DETR+PointLLM & joint-training & 3.80 & 13.84 & 1.79 & 12.75 & 25.89 & 10.12 & 0.875 & 0.612 & 0.775 & 0.384 &0.536 &0.298 & 0.394 \\
    3DETR+Qwen3 & joint-training & 4.95 & 16.31 & 2.67 & 15.49 & 36.57 & 11.28 & \textbf{0.911} & 0.622 & 0.801 & 0.395 &0.567 &0.316 & 0.418 \\
    3DETR+Florence2 & joint-training & 5.16 & \ul{18.05} & 2.58 & 15.87 & \ul{40.53} & 10.94 & 0.876 & 0.651 & \ul{0.806} & 0.421 &0.570 &0.339 & 0.433 \\
    CoDA & joint-training & \ul{5.75} & 16.87 & \ul{3.52} & \ul{18.82} & 40.21 & \ul{14.55} & 0.885 & \ul{0.680} & 0.804 & \ul{0.448} &\ul{0.602} &\ul{0.360} & \ul{0.458} \\
    \colorours \ourso & joint-training & \textbf{9.23} & \textbf{20.04} & \textbf{7.07} & \textbf{27.20} & \textbf{47.98} & \textbf{23.04} & 0.890 & \textbf{0.797} & \textbf{0.812} & \textbf{0.585} &\textbf{0.709} &\textbf{0.475} & \textbf{0.570} \\
    \specialrule{.15em}{.05em}{.05em}
    \end{tabular}
}

\label{tab:main_scannet}
\end{table*}
\begin{table*}[t]
\centering
\caption{
\textbf{Quantitative comparison on SUNRGB-D dataset.}
All models are under the joint-training setting.
The cell format, acronyms, and notations are the same as those in Tab.~\ref{tab:main_scannet}.
}
\resizebox{.95\linewidth}{!}{
    \begin{tabular}{lcccccc|ccccccc}
    \specialrule{.15em}{.05em}{.05em}
    & \multicolumn{6}{c}{Localization} & \multicolumn{7}{c}{Semantics} \\
    \cline{2-14} 
    \multirow{-2}{*}{Method} & mAP & mAP$_\text{b}$ & mAP$_\text{n}$ & Rec & Rec$_\text{b}$ & Rec$_\text{n}$ & AUC$_\text{b}$ & Cov$_\text{b}$ & AUC$_\text{n}$ & Cov$_\text{n}$ & \texttt{SS}$_\text{b}$ & \texttt{SS}$_\text{n}$ & \texttt{SS} \\ 
    \hline
    3DETR+PointLLM & 7.65 & 32.97 & 0.61 & 13.79 & 50.18 & 2.68 & \textbf{0.960} & 0.710 & 0.740 & \ul{0.328} &0.682 &\ul{0.243} & 0.554 \\
    3DETR+Qwen3 & 8.69 & \textbf{34.30} & 1.57 & 17.09 & \ul{51.55} & 7.51 & 0.954 & \ul{0.727} & 0.770 & 0.287 &\ul{0.694} &0.221 & \ul{0.557} \\
    3DETR+Florence2 & 9.21 & 33.65 & 2.42 & 17.39 & 51.20 & 7.99 & \textbf{0.960} & 0.720 & 0.764 & 0.315 &0.691 &0.241 & 0.561 \\
    CoDA & \ul{9.29} & 31.79 & \ul{3.04} & \ul{18.60} & 47.95 & \ul{10.45} & 0.944 & 0.718 & \ul{0.773} & \textbf{0.336} &0.678 &\textbf{0.260} & \ul{0.557} \\
    \colorours \ourso & \textbf{10.00} & \ul{33.98} & \textbf{3.33} & \textbf{19.97} & \textbf{52.97} & \textbf{10.80} & \ul{0.959} & \textbf{0.748} & \textbf{0.797} & 0.303 &\textbf{0.717} &0.241 & \textbf{0.579} \\
    \specialrule{.15em}{.05em}{.05em}
    \end{tabular}
}
\label{tab:main_sunrgbd}
\end{table*}

We conduct extensive experiments to validate the efficacy of our proposed \ours{} framework on the \TaskName task.
First, we explain the experimental setup, including datasets, baselines, and evaluation metrics (\secref{sec:exp_setting}).
Next, we present quantitative and qualitative comparisons with the SOTA methods on the ScanNetV2 and SUNRGB-D datasets (\secref{sec:exp_results}).
Finally, we perform ablation studies to analyze the contribution of each component in \ours{} (\secref{sec:exp_ablation}).

\subsection{Experimental setup}
\label{sec:exp_setting}
{\bf\noindent Datasets and splits.}
We evaluate \ours on two indoor 3D object detection benchmarks: ScanNetV2~\cite{dai2017scannet} and SUNRGB-D~\cite{song2015sun}.
ScanNetV2 contains 1201 training scans with 200 object classes~\cite{rozenberszki2022language}.
SUN-RGBD includes 5285 training samples covering 46 object classes.
Following~\cite{cao2023coda}, we define base/novel classes by their training-set frequency: in ScanNetV2, the 10 most frequent classes are base classes and the next 50 most frequent ones are novel classes; in SUNRGB-D, the 10 most frequent classes are base classes and the remaining 36 frequent ones are novel classes.

{\bf\noindent Baselines.}
We compare \ours with a range of open-source baselines that explore semantics via different modalities under different training settings.
Specifically, we include
(i) the SOTA open-source OV3DOD method CoDA~\cite{cao2023coda};
(ii) a 3D VLM-driven exploration baseline 3DETR+PointLLM~\cite{xu2024pointllm};
(iii) a 2D VLM-driven exploration baseline
3DETR+Florence2~\cite{xiao2024florence};
and (iv) an LLM-driven exploration baseline
3DETR+Qwen3~\cite{yang2025qwen3}.
Under the zero-shot setting, we deploy a pre-trained class-agnostic 3D object detector 3DETR~\cite{misra2021end} and query it with the corresponding explored class labels.
Under the joint-training setting, we further train the 3D detector by using the discovered class labels, following the training strategy of~\citet{cao2023coda}.
As CoDA requires user-defined vocabulary at inference, we supply the same vocabulary it uses during training.
Please refer to the supplement for the details.

{\bf\noindent Evaluation metrics.}  
As described in~\secref{sec:eval_protocol}, we evaluate the performance of all baselines and \ours{} on the \TaskName task from two complementary perspectives: localization and semantics.  
For localization, we follow~\cite{cao2023coda} and report the mAP and Average Recall (AR) at an IoU threshold of 0.25.  
For semantics, we report results using our proposed \emph{Semantic Score} (\texttt{SS}).  
To provide a more comprehensive analysis, we also report the AUC, $\operatorname{Cov}$, and \texttt{SS} for both base and novel classes.

{\bf\noindent Implementation details.}
\ours uses the class-agnostic 3D backbone 3DETR pre-trained on the base class objects as in~\citet{cao2023coda}. We train 200 epochs,
save checkpoints every 20 epochs, and select the best performing model by the validation set performance.
At inference, we exclude $\gF^{\text{E}}$ from $\gF^{\text{S}}$ for label readability. All the hyperparameters follow the default 3DETR setup~\cite{misra2021end}.
More implementation details are in the supplement.

\subsection{Experimental results}
\label{sec:exp_results}
{\bf\noindent Quantitative results.}
We compare \ourso with the baselines on ScanNetV2 quantitatively in \tabref{tab:main_scannet}.
The models jointly trained with semantic supervision consistently outperform those evaluated in the zero-shot setting across both localization and semantics metrics.
Among the jointly trained baselines, the 3D VLM-driven method with PointLLM performs the worst.
Although PointLLM is trained on large-scale 3D object datasets~\cite{xu2024pointllm, deitke2023objaverse}, its semantic predictions are often unreliable when applied to noisy or incomplete objects extracted from complex 3D scenes.

The purely language-driven method with Qwen3 produces reasonable semantics but remains inferior to the 2D VLM-driven approach using Florence2, which benefits from 2D images for more accurate semantic reasoning.
The SOTA OV3DOD method CoDA performs strongly, but it takes advantage of a pre-defined vocabulary.
In contrast, by integrating a 2D VLM-driven method with pseudo 3D proposals and feature space semantic expansion, \ours outperforms all the baselines on most localization and semantics metrics.
Quantitative results on SUNRGB-D, shown in \tabref{tab:main_sunrgbd}, show similar trends to those observed in ScanNetV2: under the joint-training setting, \ours consistently outperforms all baselines on most localization and semantics metrics.

\begin{figure*}[t]
\centering
\includegraphics[width=\linewidth]{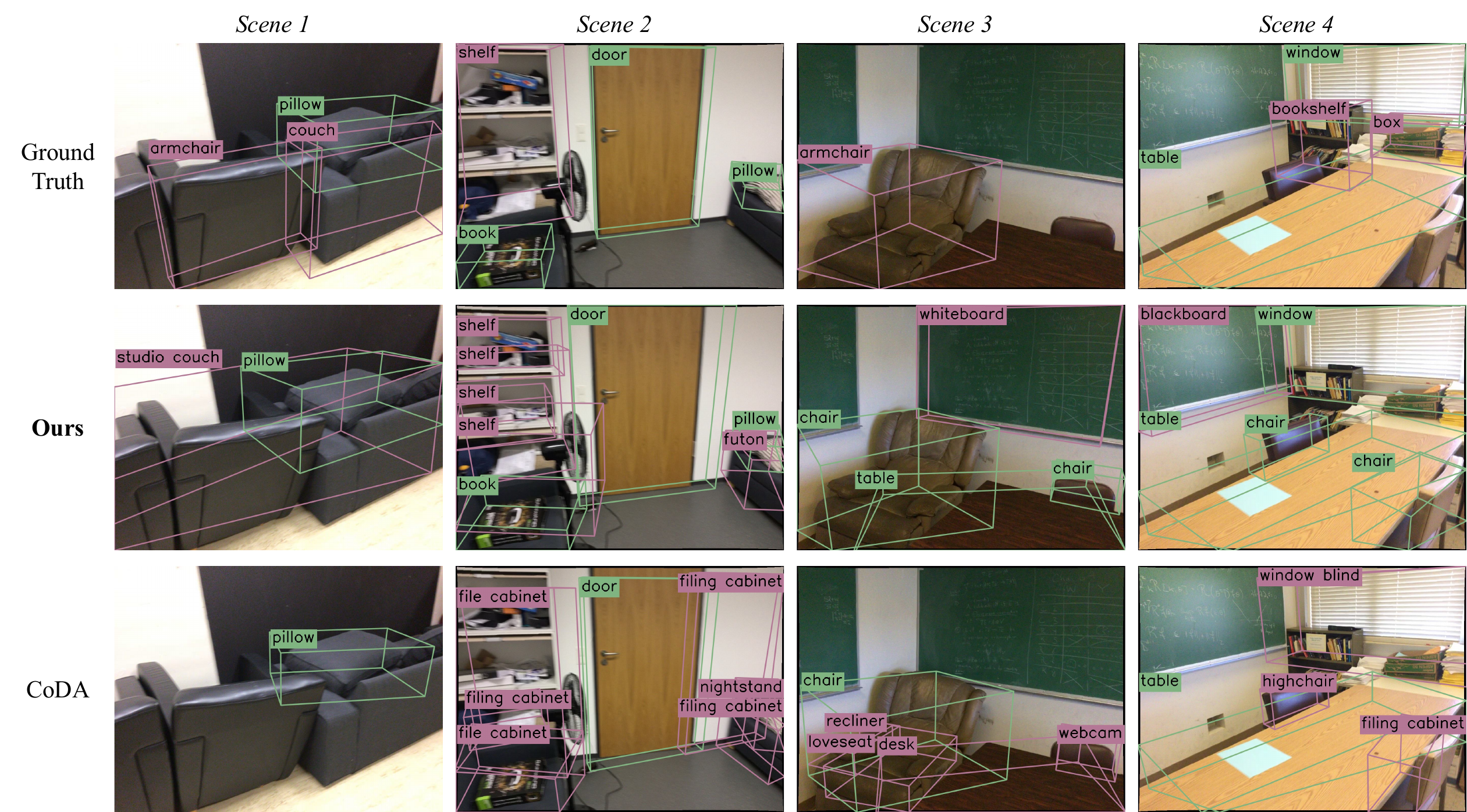}
\vspace{-1.5em}
\caption{
\textbf{Qualitative results on ScanNet validation set.}
3D bounding boxes are projected onto RGB images for visualization.
Objects from {\color{myteal}base}/{\color{mylavender}novel} classes are shown in {\color{myteal}teal}/{\color{mylavender}lavender} respectively.
}
\label{fig:qualitative_results}
\vspace{-0.35cm}
\end{figure*}
{\bf\noindent Qualitative results.}
We compare \ourso and the baseline CoDA~\cite{cao2023coda} across four sample scenes in ~\figref{fig:qualitative_results} and observe four consistent advantages:

\noindent \emph{(i) Higher coverage, including unlabeled instances.} \ours recovers extra base class objects missing from the ground truth, \eg, extra `chair' and `table' in Scene 3.
In Scene 4, it detects both an extra base class `chair' and an unannotated novel object `blackboard.'

\noindent\emph{(ii) Precise novel-class discovery.}
\ours assigns coherent class labels to novel objects.
For example, in Scene 1, \ours successfully identifies a novel-class instance `studio couch' while CoDA only recognizes the base-class object `pillow.'
In Scene 2, \ours discovers novel objects `shelf' and `futon' while CoDA predicts inaccurate labels `file cabinet' and `nightstand.'

\noindent\emph{(iii) More accurate 3D localization.}
\ours produces less noisy boxes, whereas CoDA produces imprecise geometry and fragmented detections, \eg, CoDA generated noisy boxes around `pillow' in Scene 2 and `chair' in Scene 3.

\noindent\emph{(iv) Robust base/novel disambiguation.}
When the classes are visually similar or co-occur, \ours{} resolves them correctly, \eg, \ours correctly predicts the base classes `door,' `book,' and `pillow' in Scene 2 while CoDA only recognizes `door' and recognizes the other two as novel objects.
In Scene 4, \ours correctly identifies base object `window' while CoDA recognizes it as a novel class.

Overall, \ours{} exhibits stronger semantic reasoning across both base and novel classes while maintaining precise 3D localization.

\begin{table}[t]
\centering
\caption{
\textbf{Ablation studies of \ours's components.} Acronyms: VSG: VLM semantics generation; PBG: pseudo boxes generation; FSSE: feature space semantics expansion.
}
\vspace{-.5em}
\resizebox{0.9\linewidth}{!}{
    \begin{tabular}{ccccc|c}
    \specialrule{.15em}{.05em}{.05em}
    & & & \multicolumn{2}{c}{Localization} & Semantics \\
    \cline{4-6}
    \multirow{-2}{*}{VSG} &  \multirow{-2}{*}{PBG} & \multirow{-2}{*}{FSSE} & mAP & Recall & \texttt{SS} \\ 
    \hline
    \ccheck & \ccross & \ccross & 5.16 & 15.87 & 0.433 \\
    \ccheck & \ccheck & \ccross & 8.65 & 25.44 & 0.552 \\
    \colorours\ccheck & \ccheck & \ccheck & \textbf{9.23} & \textbf{27.20} & \textbf{0.570} \\
    \specialrule{.15em}{.05em}{.05em}
    \end{tabular}
}
\label{tab:main_ablation}
\end{table}
\subsection{Ablation studies}
\label{sec:exp_ablation}
We ablate each of the proposed components in \ours, with results shown in ~\tabref{tab:main_ablation}.
Starting from the baseline that only utilizes semantics discovered from VLM captions, the model achieves 5.16 mAP and 0.433 in \texttt{SS}. 
Incorporating 3D pseudo proposals for additional supervision significantly 
improves both the localization and semantics performance, boosting mAP from 5.16 to 8.65 and \texttt{SS} from 0.433 to 0.552.
This shows that pseudo 3D proposals provide effective supervision for transferring 2D semantic knowledge into 3D.
Finally, applying the feature space semantics expansion (FSSE) strategy further enhances performance, reaching 9.23 mAP and 0.570 in \texttt{SS}.
This verifies that FSSE enables the model to explore broader and more diverse semantics beyond explicit text.
Overall, integrating all the components reaches the best performance across both localization and semantic metrics.
\section{Conclusion}
\label{sec:concl}
We propose {\bf Auto-vocabulary 3D object detection} (AV3DOD), a new task that moves beyond the user-defined semantics toward autonomous semantic discovery in 3D scenes.  
To support AV3DOD, we propose the \textit{Semantic Score} metric to evaluate semantic quality.
We propose \ours, a framework that leverages 2D vision-language models to generate rich semantic candidates through image captioning, pseudo 3D box generation, and feature space semantics expansion.
Our comprehensive experiments on ScanNetV2 and SUNRGB-D confirm that \ours vastly enhances both localization and semantic understanding compared to the SOTA methods.
We hope that this new task and \ours inspire further research on open-world 3D understanding and autonomous semantic discovery.

\clearpage
\setcounter{page}{1}

\setcounter{section}{0}
\renewcommand{\theHsection}{A\arabic{section}}
\renewcommand{\thesection}{A\arabic{section}}
\renewcommand{\thetable}{A\arabic{table}}
\setcounter{table}{0}
\setcounter{figure}{0}
\renewcommand{\thetable}{A\arabic{table}}
\renewcommand\thefigure{A\arabic{figure}}
\renewcommand{\theHtable}{A.Tab.\arabic{table}}
\renewcommand{\theHfigure}{A.Abb.\arabic{figure}}
\renewcommand\theequation{A\arabic{equation}}
\renewcommand{\theHequation}{A.Abb.\arabic{equation}}

\section*{Appendix}
\noindent The appendix is organized as follows:
\begin{itemize}
\item In~\secref{sec:eval_metrics}, we provide details of the proposed evaluation protocol.
\item In~\secref{sec:implement_ours}, we describe additional implementation details of \ours{}.
\item In~\secref{sec:implement_baselines}, we describe implementation details of baselines.
\item In~\secref{sec:hyper_search}, we document hyperparameter search details for our proposed FSSE.
\item In~\secref{sec:add_qualitative}, we report additional qualitative examples.
\item In~\secref{sec:compare_other_auto}, we provide comparison with other auto-vocabulary work.
\item In~\secref{sec:computational_cost}, we report the computational cost of our model.
\end{itemize}

\section{Evaluation Metric Details}
\label{sec:eval_metrics}
We provide additional details on the proposed semantic score (\texttt{SS}) defined in \equref{eq:ss}.

{\bf\noindent Matching protocol.}
For each ground-truth box, we identify its matched prediction as the one with the highest Intersection-over-Union (IoU) exceeding $0.25$, following standard practice~\cite{cao2023coda}.
Each ground-truth box matches at most one prediction, and unmatched ground-truth boxes are excluded from the semantic evaluation.

{\bf\noindent Text embedding generation.}
To obtain class embeddings, we follow prior work~\cite{cao2023coda} and use the CLIP ViT-B/16~\cite{radford2021learning} text encoder with the prompt template 
\textit{``a photo of a \{OBJECT\} in the scene''}, where \textit{`\{OBJECT\}'} denotes the class label.
All text embeddings are L2-normalized.

{\bf\noindent AUC computation.}
A prediction is considered semantically correct if its cosine similarity with the ground-truth embedding exceeds a threshold $\tau$.
We evaluate accuracy over a fixed set of thresholds
\[
\tau \in \{0.0, 0.2, 0.4, 0.6, 0.8, 1.0\},
\]
and compute the area under this accuracy-threshold curve using the trapezoidal rule.
We report $\operatorname{AUC}$ separately for base and novel classes.

{\bf\noindent Class balancing.}
The final semantic score \texttt{SS} aggregates the base class and novel class components using weights $(\alpha_{\text{b}}, \alpha_{\text{n}})$ proportional to the number of ground-truth objects in each group.
For the ScanNet dataset, $(\alpha_{\text{b}}, \alpha_{\text{n}}) = (0.405,\, 0.595)$.
For the SUNRGB-D dataset, $(\alpha_{\text{b}}, \alpha_{\text{n}}) = (0.710,\, 0.290)$.
These values are computed from the full validation set annotations.

\section{Implementation Details of \ours{}}
\label{sec:implement_ours}
We provide additional implementation details of \ours{}.
{\bf\noindent Pseudo box generation.}
For the 2D VLM $F_{\text{VLM}}^{\text D}$, we use \texttt{Florence-2-large} with the task prompt \textit{`OD'} to locate the 2D bounding boxes and predict class labels.
Then we query the same model on each generated 2D box with the task prompt \textit{`REGION\_TO\_SEGMENTATION'} to obtain the object segmentation mask.
For estimating the object orientation via PCA, we follow the implementation of \citet{lu2023open}.

{\bf\noindent Feature set sizes.}
For the ScanNet dataset, the base class features $\gF^{\text B}$ contain 10 entries.
The VLM caption features $\gF^{\text C}$ contribute 658 candidates, and the pseudo-label features $\gF^{\text P}$ add 210 classes, yielding 772 non-overlapping features in $\gF^{\text{Base}}$.
We generate an additional 30\% (\ie 231) expanded features $\gF^{\text E}$ through feature space semantics expansion, 
resulting in a unified vocabulary $\gF^{\text{S}}$ of size 1003.
For the SUNRGB-D dataset, the sizes of $\gF^{\text B}$, $\gF^{\text C}$, $\gF^{\text P}$, $\gF^{\text{Base}}$, $\gF^{\text E}$, and $\gF^{\text{S}}$ are 10, 434, 220, 559, 111, and 670 respectively.
Complete lists for both datasets will be provided with the released code.

{\bf\noindent Training details.}
We follow the default 3DETR setup used in \citet{misra2021end} and \citet{cao2023coda} for experiments on both datasets.
Specifically, we set the object localization loss weights $\lambda_{\text{obj}}$, $\lambda_{\text{center}}$, $\lambda_{\text{size}}$ and  $\lambda_{\text{angle}}$ in \equref{eq:loss_loc} to 1.0, 0.5, 1.0, 0.1 respectively.
Both semantic loss weights $\lambda_{\text{d}}$ and $\lambda_{\text{s}}$ in \equref{eq:loss_final} are set to 1.
For the image encoder $E^{\text I}$ and text encoder $E^{\text T}$, we use a frozen pre-trained CLIP with ViT-B/16.
To prevent degradation in base-class performance due to pseudo-box supervision, we upweight the annotated ground-truth boxes by duplicating them during training.
For ScanNet, each ground-truth box is included three times, and for SUNRGB-D, each box is repeated twice.

\begin{table*}[t]
\centering
\caption{
\textbf{Hyperparameter search for FSSE on the ScanNet dataset.}
The number of expanded samples $K$ is expressed as a proportion of the full base feature set $\gF^{\text{Base}}$.
The cell format, acronyms, and notations are the same as those in Tab.~\ref{tab:main_scannet}.
Results of the open-vocabulary baseline CoDA~\cite{cao2023coda} are included for reference.
}
\resizebox{1\linewidth}{!}{
    \begin{tabular}{cccccccc|ccccccc}
    \specialrule{.15em}{.05em}{.05em}
    &  & \multicolumn{6}{c}{Localization} & \multicolumn{7}{c}{Semantics} \\
    \cline{3-15} 
    \multirow{-2}{*}{$K$} & \multirow{-2}{*}{($\theta^{E}_{\text{min}}$, $\theta^{E}_{\text{max}}$)} & mAP & mAP$_\text{b}$ & mAP$_\text{n}$ & Rec & Rec$_\text{b}$ & Rec$_\text{n}$ & AUC$_\text{b}$ & Cov$_\text{b}$ & AUC$_\text{n}$ & Cov$_\text{n}$ & \texttt{SS}$_\text{b}$ & \texttt{SS}$_\text{n}$ & \texttt{SS} \\ 
    \hline
    10\% & (0.7, 0.8) & 8.95 & 19.85 & 6.76 & \ul{27.23} & 47.19 & \ul{23.24} & \ul{0.892} & 0.773 & \textbf{0.816} & 0.566 & 0.690 & 0.462 & 0.554 \\
    10\% & (0.6, 0.9) & 9.02 & 19.18 & \ul{6.99} & 26.85 & 47.39 & 22.74 & 0.888 & 0.789 & 0.811 & 0.565 & 0.701 & 0.458 & 0.557 \\
    20\% & (0.7, 0.8) & 8.60 & 19.28 & 6.46 & 26.13 & 44.96 & 22.37 & 0.884 & 0.773 & \ul{0.814} & 0.537 & 0.683 & 0.437 & 0.536 \\
    20\% & (0.6, 0.9) & \ul{9.01} & \ul{19.89} & 6.83 & 26.22 & \textbf{48.81} & 21.70 & \textbf{0.898} & 0.776 & 0.808 & 0.582 & 0.697 & 0.470 & 0.562 \\
    \colorours 30\% & (0.7, 0.8) & \textbf{9.23} & \textbf{20.04} & \textbf{7.07} & 27.20 & \ul{47.98} & 23.04 & 0.890 & \ul{0.797} & 0.812 & \ul{0.585} & \ul{0.709} & \ul{0.475} & \ul{0.570} \\
    30\% & (0.6, 0.9) & 8.94 & 19.04 & 6.92 & \textbf{28.16} & 47.86 & \textbf{24.22} & 0.890 & \textbf{0.806} & 0.811 & \textbf{0.601} & \textbf{0.717} & \textbf{0.487} & \textbf{0.580} \\
    \hline
     \multicolumn{2}{l}{CoDA~\cite{cao2023coda}} & 5.75 & 16.87 & 3.52 & 18.82 & 40.21 & 14.55 & 0.885 & 0.680 & 0.804 & 0.448 & 0.602 & 0.360 & 0.458 \\
    \specialrule{.15em}{.05em}{.05em}
    \end{tabular}
}

\label{tab:supp_expansion_hyper}
\end{table*}

\section{Implementation Details of Baselines}
\label{sec:implement_baselines}
We provide implementation details of the baselines that discover semantics from different modalities on the ScanNet dataset.

{\bf\noindent 2D VLM-driven baseline.}
We use the same procedure for generating class candidates as the VLM captioning process in \ours.
Each scene is queried using \texttt{Florence-2-large} with the task prompt \textit{`CAPTION'}.
We then apply the NLP toolkit spaCy \cite{spaCy} to parse the generated caption and extract all nouns.
We then aggregate the unique nouns across all scenes to form the discovered vocabulary, yielding a total of 658 distinct labels.

{\bf\noindent 3D VLM-driven baseline.}
To explore semantics directly in the 3D domain, we query the SOTA 3D object captioner \texttt{PointLLM\_7B\_v1.2}~\cite{xu2024pointllm}.
We first apply a pretrained detector 3DETR~\cite{misra2021end} to the training data to obtain class-agnostic 3D bounding boxes for each scene.
Each detected box is then passed to PointLLM with the prompt: \textit{"What is this?"}, following the protocol from~\citet{xu2024pointllm}.
As PointLLM outputs sentence-level descriptions, we further apply a language model Qwen3~\cite{yang2025qwen3} to generate a one-word description for each box.
Specifically, we query \texttt{Qwen3-30B-A3B-Instruct-2507} model with the prompt: \textit{``I am going to describe a 3D object. Please summarize the object category using one word. Your answer should only contain that one word without any additional explanation.''}
Finally, we aggregate all labels and choose the most frequent 650 words to match the vocabulary size of the 2D VLM-driven baseline.

{\bf\noindent LLM-driven baseline.}
To explore semantics purely from language, we query the \texttt{Qwen3-Max-Instruct} model~\cite{yang2025qwen3} with the prompt: \textit{``Here are 10 objects that appear in common indoor scenes: $\{\text{base classes}\}$. Please generate a list of possible vocabulary that is related to these objects. Please return a list with 650 categories.''},
where we provide the base class labels in \textit{``$\{\text{base classes}\}$''}.
The full list of the generated vocabularies of different baselines on both datasets will be provided in the released code.

\section{Hyperparameter Search}
\label{sec:hyper_search}
We provide details on the hyperparameter search conducted for our proposed feature space semantics expansion (FSSE).
Specifically, we search over different combinations of number of samples $K$, the threshold lower bound $\theta^{E}_{\text{min}}$ and upper bound $\theta^{E}_{\text{max}}$.
The results are shown in \tabref{tab:supp_expansion_hyper}.
Our proposed \ours{} consistently outperforms the baseline CoDA across all configurations, demonstrating the robustness of FSSE.
We find that setting $K$ to 30\% of the base feature set size yields stronger performance in both localization and semantics metrics compared to lower ratios.
We select the optimal $K$, $\theta^{E}_{\text{min}}$ and $\theta^{E}_{\text{max}}$ based on overall mAP.

\begin{figure*}[t]
\centering
\includegraphics[width=\linewidth]{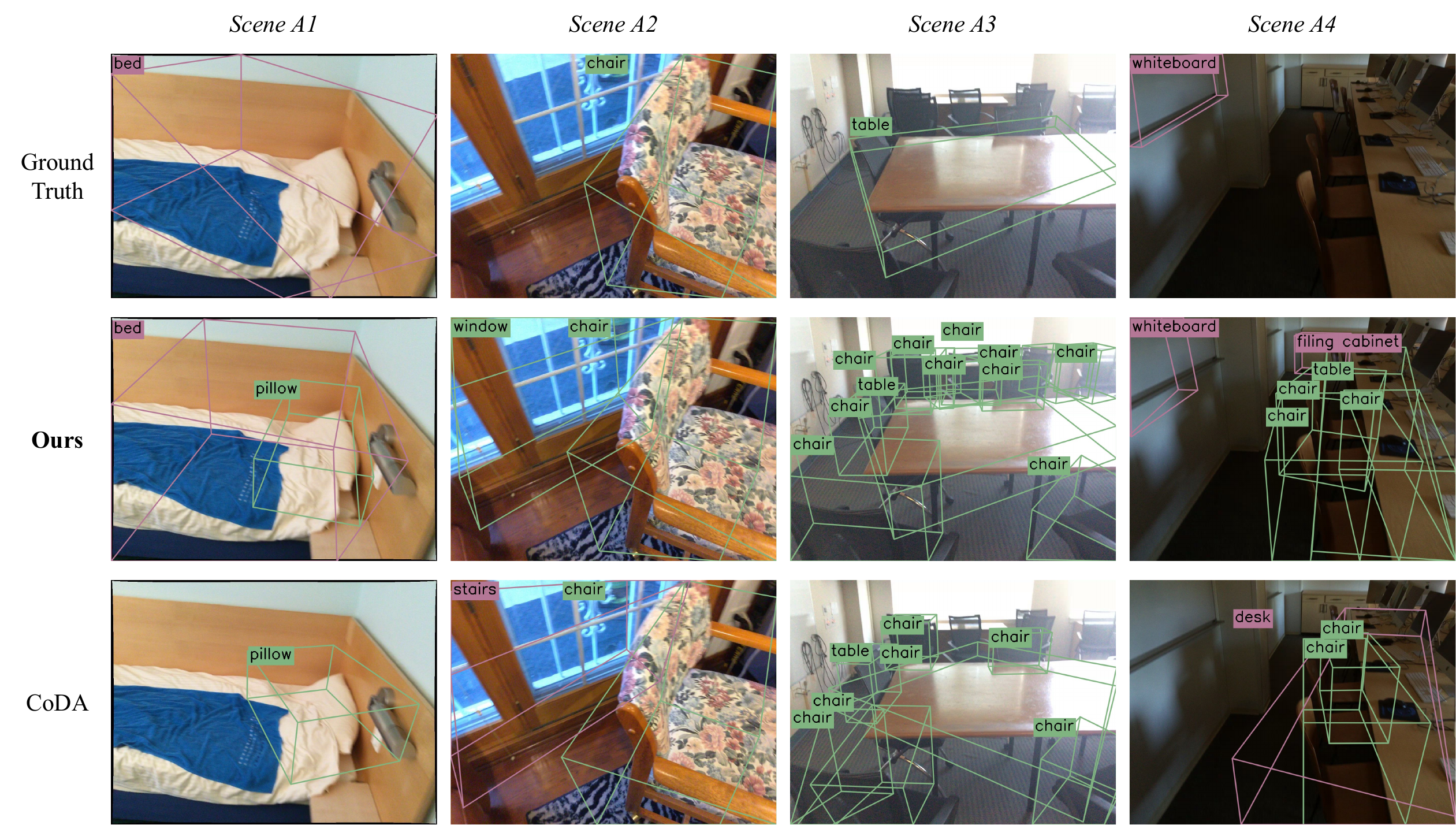}
\caption{
\textbf{Additional qualitative results on the ScanNet validation set.}
3D bounding boxes are projected onto RGB images for visualization.
Objects from {\color{myteal}base}/{\color{mylavender}novel} classes are shown in {\color{myteal}teal}/{\color{mylavender}lavender} respectively.
}
\label{fig:supp_qualitative}
\end{figure*}
\section{Additional Qualitative Examples}
\label{sec:add_qualitative}
\figref{fig:supp_qualitative} presents additional qualitative examples, which consistently illustrate the advantages of \ours{} over CoDA~\cite{cao2023coda}, as discussed in \secref{sec:exp_results}:

\noindent \emph{(i) Higher coverage, including unlabeled instances.} \ours recovers additional base class objects missing from the ground truth.
For instance, in Scene A1, it detects an unannotated base-class `pillow'.
In Scene A2, it correctly identifies an additional `window' that CoDA misclassifies as `stairs'.

\noindent\emph{(ii) Precise novel-class discovery.}
\ours assigns coherent class labels to novel objects.
It successfully identifies the novel class `bed' in Scene A1 and `whiteboard' in Scene A4, both of which CoDA fails to recognize.

\noindent\emph{(iii) More accurate 3D localization.}
Compared to CoDA, \ours{} generates better-aligned bounding boxes.
For example, CoDA's `table' in Scene A3 and `desk' in Scene A4 exhibit imprecise geometry, while \ours{} produces boxes better aligned with object boundaries.

\noindent\emph{(iv) Robust base/novel disambiguation.}
When visually similar base and novel objects co-occur, \ours{} distinguishes them more reliably.
In Scene A4, it correctly recognizes the base class object `table', whereas CoDA misclassifies it as some novel class `desk'.

These examples further demonstrate that \ours{} can accurately localize objects and generate reliable labels for both base and novel classes.

\section{Other Auto-vocabulary Work}
\label{sec:compare_other_auto}
We review related approaches on semantic discovery in prior and concurrent auto-vocabulary work.

{\bf\noindent 2D auto-vocabulary methods.}
Recent methods such as \citet{conti2023vocabulary}, \citet{ulger2025auto} and \citet{conti2024vocabulary} leverage 2D VLMs to construct candidate vocabularies for 2D image classification or segmentation.
However, these approaches operate solely on 2D images and cannot be directly applied to our setting, where the input is a 3D point cloud.
For comparison, we therefore introduce a 2D VLM-driven baseline that extracts semantics from the 2D image corresponding to each 3D scene.

{\bf\noindent 3D auto-vocabulary methods.}
\citet{mei2025vocabulary} also uses a 2D VLM to generate the semantics from an image corresponding to the 3D scene, making it conceptually aligned with our 2D VLM–driven baseline.
\citet{wei20253d} distills a 2D captioning model into a 3D point cloud captioner, enabling scene-level caption generation at inference time.
Despite our best efforts, we were unable to fully reproduce their results due to the partially released code base.
As an alternative, we include a 3D VLM-driven baseline which utilizes the SOTA 3D captioner to directly generate candidate classes.
We will provide a more comprehensive comparison once the full implementation becomes publicly available.

\begin{table}[t]
\centering
\caption{\textbf{Computational cost for the proposed \ours.}
We compare with the SOTA open-vocabulary model CoDA~\cite{cao2023coda}.
}
\resizebox{1\linewidth}{!}{
    \begin{tabular}{lcc}
    \specialrule{.15em}{.05em}{.05em}
     Method & \# Trainable Parameters (M) & Inference Time (s) \\
    \hline
     CoDA~\cite{cao2023coda} & 24.6 & 0.0674 \\
    \colorours \ours & 24.6 & 0.0673 \\
    \specialrule{.15em}{.05em}{.05em}
    \end{tabular}
}
\label{tab:supp_computation}
\end{table}
\section{Computational Cost}
\label{sec:computational_cost}
We report the number of trainable parameters and inference time of \ours{} and the baseline model CoDA in~\tabref{tab:supp_computation}. 
The experiments are conducted on the ScanNet validation set using a single NVIDIA L40S GPU.
As \ours employs the same 3DETR backbone~\cite{misra2021end} as CoDA and constructs the super vocabulary features $\gF^{\text S}$ entirely during training, it introduces no additional computational overhead during inference.
\ours{} matches CoDA in the trainable parameter count and inference speed, while achieving superior localization and semantic performance without extra runtime cost.

\clearpage
{
    \small
    \bibliographystyle{ieeenat_fullname}
    \bibliography{main}
}

\end{document}